\documentclass[10pt,twocolumn,letterpaper]{article}

\usepackage{cvpr}              

\usepackage[dvipsnames]{xcolor}

\definecolor{cvprblue}{rgb}{0.21,0.49,0.74}
\usepackage[pagebackref,breaklinks,colorlinks,citecolor=cvprblue]{hyperref}

\usepackage{cite}
\usepackage{amsmath,amssymb,amsfonts}
\usepackage{algorithmic}
\usepackage{graphicx}
\usepackage{textcomp}
\usepackage{xcolor}
\usepackage{xspace}
\usepackage{url}
\usepackage{verbatim}
\usepackage{tabularx}
\usepackage{multirow}
\usepackage{graphicx}
\usepackage{textcomp}
\usepackage{import}
\usepackage{array,multirow,graphicx}
\usepackage{float}
\usepackage{caption}
\usepackage{subcaption}
\usepackage{booktabs}

\title{\bf Black-box Adversarial Attacks on CNN-based SLAM Algorithms}

\author{Maria Rafaela Gkeka\textsuperscript{1}, Bowen Sun\textsuperscript{2}, Evgenia Smirni\textsuperscript{2}, Christos D. Antonopoulos\textsuperscript{1}, \\Spyros Lalis\textsuperscript{1}, Nikolaos Bellas\textsuperscript{1} \\ \\
\textsuperscript{1}Department of Electrical and Computer Engineering, \\
        University of Thessaly, Greece\\ 
        {\tt\small margkeka@uth.gr, cda@uth.gr, lalis@uth.gr, nbellas@uth.gr} \\
  \textsuperscript{2}Department of Computer Science, William \& Mary, USA\\
        {\tt\small bsun02@wm.edu, esmirni@cs.wm.edu}\\
}

\begin{document}
\maketitle
\begin{abstract}
Continuous advancements in deep learning have led to significant progress in feature detection, resulting in enhanced accuracy in tasks like Simultaneous Localization and Mapping (SLAM). Nevertheless, the vulnerability of deep neural networks to adversarial attacks remains a challenge for their reliable deployment in applications, such as navigation of autonomous agents.
Even though CNN-based SLAM algorithms are a growing area of research
there is a notable absence of a comprehensive presentation and examination of adversarial attacks targeting CNN-based feature detectors, as part of a SLAM system. 
Our work introduces black-box adversarial perturbations applied to the RGB images fed into the GCN-SLAM algorithm. Our findings on the TUM dataset~\cite{sturm2012benchmark} reveal that even attacks of moderate scale can lead to tracking failure in as many as $76\%$ of the frames. 
Moreover, our experiments highlight the catastrophic impact of attacking depth instead of RGB input images on the SLAM system.
\end{abstract}    
\section{Introduction}
\label{sec:intro}

The proliferation of autonomous robots, unmanned aerial vehicles (UAVs), and driverless cars has generated a demand for the creation of precise maps of the environments they perceive, along with the necessity to monitor the positions and trajectories of these agents within these maps. Simultaneous Localization and Mapping (SLAM) algorithms have been used to solve these problems by integrating data from a diverse array of sensors, including stereo/mono and RGB-D cameras, laser scanners (lidars), and Inertial Measurement Units (IMUs). In such power-constrained platforms, SLAM implementations typically use sparse SLAM algorithms. These algorithms diminish computational demands by retaining only a sparse subset of essential feature points, often limited to the sole purpose of localization. 

Traditional feature detectors such as ORB (Oriented FAST and Rotated BRIEF) are integral to sparse SLAM, as they enable the system to efficiently handle the limited computational resources by tracking distinctive visual landmarks in an image.
These landmarks serve as the sparse set of key points for agent localization. 
State-of-the-art sparse SLAM algorithms such as ORB-SLAM2~\cite{mur2017orb} focus on using the most salient key points 
of an image selected from multiple previous frames, to reduce the impact of noise, scale, rotation, and lighting conditions and subsequently enhance their robustness against environmental variations. 

Recent, state-of-the-art feature detectors leverage the advances in deep neural networks (and, specifically, in visual transformers~\cite{KhanNHZKS22_VisionTransf})
 to establish accurate topological associations between image pairs to enable robust and precise feature matching and tracking~\cite{tang2019gcnv2, detone2018superpoint, SarlinDMR20, sun2021loftr, WangZYPS22Matchformer, ChenLZTZFMTQ22ASpanFormer, LightGlue}. Although deep neural networks achieve superior matching accuracy, they are susceptible to adversarial attacks 
based on even minor pixel perturbations of an original image~\cite{fgsm2014goodfellow, carlini2017towards}. An increasing body of research addresses this issue for image classification~\cite{YuanHZL19AdvExamples}, semantic segmentation~\cite{ArnabMT18SemSegm}, and pose estimation~\cite{chawla2022adversarial}. 

Our work proposes effective adversarial attack strategies tailored specifically for neural networks (NNs) for feature detection 
that are used as a front-end in a SLAM pipeline.  Given their considerable robustness, feature detectors present unique challenges in attacking them compared to neural networks for image classification or object detection. 
Our approach is designed for the GCNv2 feature detector~\cite{tang2019gcnv2}, although it also applies to other feature detectors (see section~\ref{subsec:transferability}).

Past work on generating adversarial attacks focuses on white-box scenarios where the attacker has complete knowledge and access to the architecture of the target neural network, as well as its parameters and training data~\cite{WangLCRW22WhiteBoxAttack}. Yet, in most realistic settings, an attacker cannot assume knowledge of or have access to the underlying feature detector. Here, we assume a black-box approach where the attacker changes the expected output of the feature detection network by tampering only with its input~\cite{papernot2017practical} and without having access to the network internals.

\begin{figure*}[h]
  \centering    
  \includegraphics[width=0.9\textwidth]{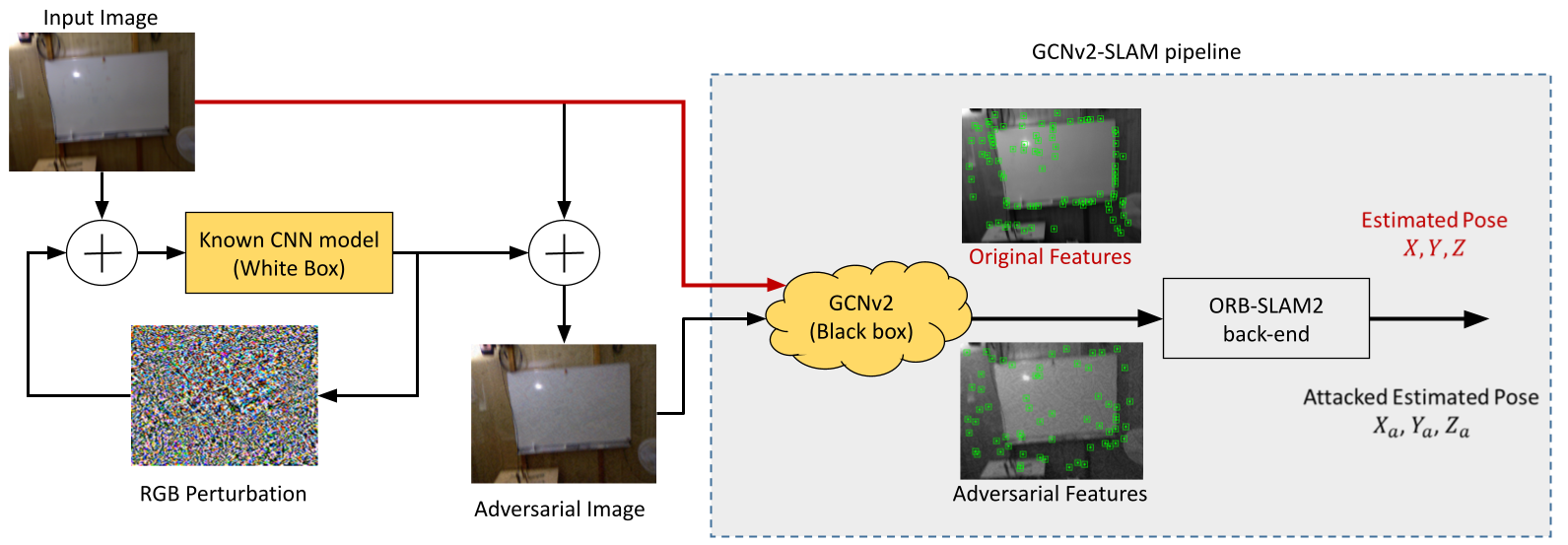}
  \caption{Block diagram of our approach to black-box attacks on CNN-based feature detectors and the SLAM pipeline (path with black arrows). The path with red arrows is the baseline (i.e. no attacks) 
  GCN-SLAM pipeline.}
  \label{fig:blackbox_bd}
\end{figure*} 

We demonstrate that a SLAM system is vulnerable to black-box adversarial attacks. 
The attack on the known model (in our case, a CNN) results in perturbations added to the original frames which are fed to the feature detection network. Our experiments use the TUM dataset~\cite{sturm2012benchmark} and demonstrate that the GCNv2 feature detector is significantly vulnerable to both targeted and untargeted attacks, 
driving the GCN-SLAM pipeline to lose its ability to track frames and accurately capture the output trajectory.
We obtain similar results when the attack is not applied to the whole frame but is spatially limited to the objects detected using an object detection algorithm such as YOLOv4~\cite{bochkovskiy2020yolov4}. 


Our contributions are summarized as follows:
\begin{itemize}
\item{We introduce and experimentally evaluate attacks targeting neural network-based feature detectors (specifically, GCNv2), based on various strategies such as FGSM~\cite{fgsm2014goodfellow} and PGD~\cite{pgd2017madry}. 
This is the first work to propose and study adversarial attacks on NN-based feature detectors.}
\item{We evaluate the effects of black-box adversarial attacks on the SLAM pipeline, focusing on its ability to accurately estimate output trajectories without large errors compared to ground truth. We assess the robustness of GCN-SLAM~\cite{tang2019gcnv2} using well-established CNNs including InceptionResNetV2~\cite{SzegedyIVA17_InceptionResnet}, MobileNetV2~\cite{sandler2018mobilenetv2} and DenseNet~\cite{huang2017densely} to generate adversarial images.}
\item{We design and evaluate adversarial attacks that are launched only to (i) a subset of frames of the trajectory, and to (ii) image regions that contain specific objects.}
\end{itemize}

The remainder of this paper is organized as follows. Section~\ref{sec:PreviousWork} discusses related work. Section~\ref{sec:methodology} introduces the methodology of the proposed adversarial attacks, and section~\ref{sec:experiments}  presents and discusses the results of our experimental evaluation. Finally, Section~\ref{sec:concl} concludes the paper.


\section{Related Work}
\label{sec:PreviousWork}

Most existing research on adversarial attacks targets models for image classification and object detection. Our research is the first to consider adversarial perturbations targeted at CNN-based feature detectors. 
In this section, we present several previous studies relevant to our work.

\textbf{Black-box attacks}.  Several studies propose adversarial attacks using black-box techniques~\cite{guo2019simple, ilyas2018black, papernot2017practical}. In medical imaging, Finlayson et al. show how to attack medical deep learning classifiers by experimenting with white-box and black-box patches and PGD attacks on three types of illnesses~\cite{finlayson2018adversarial}. DPATCH is a black-box adversarial patch-based attack for mainstream object detectors, which fools network predictions by performing targeted and untargeted attacks to the bounding box regression and object classification~\cite{Liu2018DPATCHAA}. 
Black-box adversarial attacks have been deployed for object detection~\cite{Haoran2021ACA, lapid2023patch}, video recognition~\cite{jiang2019black, Xie2021Universal3P}, and speech emotional recognition~\cite{Gao2022BlackboxAA}.
In contrast, our work explores targeted and untargeted FGSM and PGD black-box attacks
to an entirely distinct domain, a SLAM system.

\textbf{White-box attacks}. In a recent paper Nemcovsky et al. explored visual odometry (VO) methods by investigating the impact of patch adversarial attacks on a learning-based VO model~\cite{nemcovsky2022physical}. The study involves testing the model using both synthetic and real data captured from a drone operating in an indoor arena. Chawla et al. studies adversarial PGD attacks on pose estimation networks and tests the transferability of perturbations to other networks~\cite{chawla2022adversarial}. Pose estimation networks play an important role in visual odometry (VO) algorithms, as they are responsible for predicting the camera pose based on image pairs. In contrast, SLAM is designed to achieve a globally consistent estimation of the camera trajectory and map, where visual odometry (VO) is just one of the components within the broader SLAM framework. Our work is different from~\cite{chawla2022adversarial} in two primary aspects: (a) it involves applying black-box attacks and (b) it addresses a more complex application compared to VO.

 Yao et al. proposes an  attack called ATI-FGSM~\cite{yao2020miss} that targets
 the precise detection of reference points in medical images.~\cite{abdelfattah2021adversarial} attacks networks that take as input both LiDAR images and point clouds to detect 3D cars, by rendering an adversarial texture on a 3D object. Finally,~\cite{im2022adversarial} highlights the severe vulnerability of the  YOLOv4~\cite{bochkovskiy2020yolov4} object detector after experimenting with white-box FGSM and PGD attacks on an autonomous driving image dataset.

\section{Methodology}
\label{sec:methodology}

In this section, we discuss the development of black-box adversarial attacks on a CNN-based feature detector. 

\subsection{Black-box VS White-box attacks}
An adversarial attack is a technique that uses deliberately crafted, often imperceptible perturbations to input data to manipulate or deceive a machine learning model. 
Adversarial attacks can be categorized into two groups based on the attacker's level of knowledge about the target model architecture, parameters, and training data. In white-box attacks, the attacker has complete knowledge of the network details, while in black-box attacks (followed in our work), the attacker's understanding of the model (other than its input/output behavior) is restricted or even completely absent. 

Fig.~\ref{fig:blackbox_bd} shows a schematic overview of the proposed attack.
We focus on adversarial attacks 
which require access to gradient information from the model's loss function to generate adversarial images that feed into the GCN-SLAM pipeline. Since we do not have access to the target network's (GCNv2) loss function, we generate the adversarial images using a known CNN model, which is a pre-trained InceptionResNetV2 model for image classification (``Known CNN model'' in Fig.~\ref{fig:blackbox_bd}). As we explain in section~\ref{sec:experiments}, the methodology is not restricted by the selection of the known network.
Leveraging this known CNN model, we generate adversarial perturbations for the RGB input images of the TUM dataset, employing attack methods like FGSM~\cite{fgsm2014goodfellow} or PGD~\cite{pgd2017madry}. 
We use the system error after the attack compared to the baseline path (red arrows in Fig.~\ref{fig:blackbox_bd}) as the metric to assess the success of the attack. 

GCN-SLAM consists of the GCNv2 (the CNN-based feature detector) and the ORB-SLAM2 back-end, a popular SLAM algorithm known for its robustness and performance in various real-world scenarios~\cite{mur2017orb}. 
ORB-SLAM2 back-end makes our work challenging since we are not just attacking 
an isolated CNN-based network, but we are investigating how an 
attack can affect a real system where the CNN network is just part of a more complex pipeline. Notably, ORB-SLAM2 employs several techniques to  
ensure that 
it continues to perform well even in complex scenes, 
including:
\textit{Loop closure} detection that helps correct drift error by detecting and closing loops in the trajectory, \textit{bundle adjustment} that enhances map consistency by optimizing the camera poses and map points globally, and \textit{re-localization} which -- in case of tracking failure -- attempts to recognize the current location within the map by matching keyframe descriptors with the map, helping the system regain its position and orientation.

\subsection{Attack methods}

In machine learning, adversarial attacks are typically classified into \textit{untargeted} and \textit{targeted}, each with its unique set of objectives. Untargeted attacks aim to generate adversarial examples that induce the misclassification of input data by a machine learning model, without specifying a particular target class. 
In contrast, targeted attacks are driven by specific goals. In these attacks, the assailant produces adversarial examples that coerce the model into misclassifying input data as a preselected, specific target class of their choosing.

In this work, we conduct experiments involving both untargeted and targeted attacks and assess their effect on the ability of SLAM to continue tracking the movement of an agent. We apply two types of attack methods to CNNs, FGSM and PGD, described below.

\begin{figure*}[bt]
     \centering
     \begin{subfigure}[b]{0.45\columnwidth}
         \centering
         \includegraphics[width=\columnwidth]{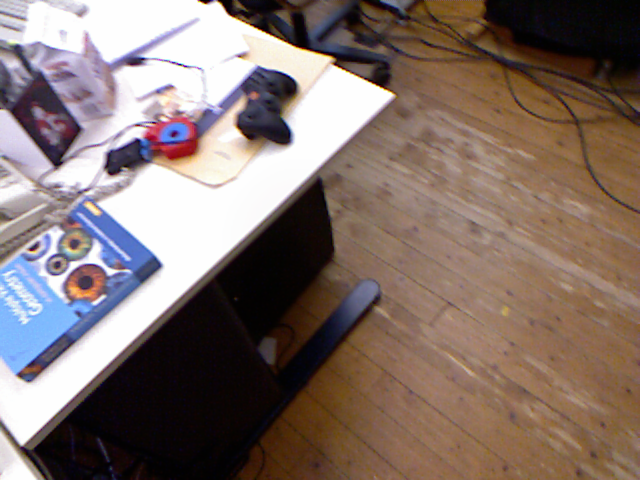}
         \caption{Original image}
     \end{subfigure}
     \hfill
     \begin{subfigure}[b]{0.45\columnwidth}
         \centering
         \includegraphics[width=\columnwidth]{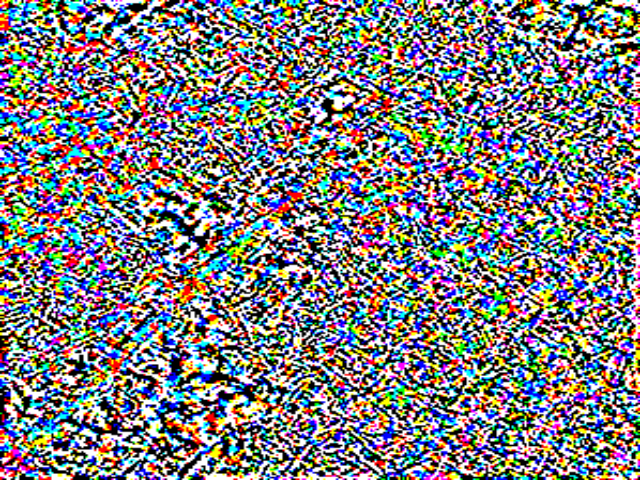}
         \caption{Adversarial perturbation}
     \end{subfigure}
     \hfill
     \begin{subfigure}[b]{0.45\columnwidth}
         \centering
         \includegraphics[width=\columnwidth]{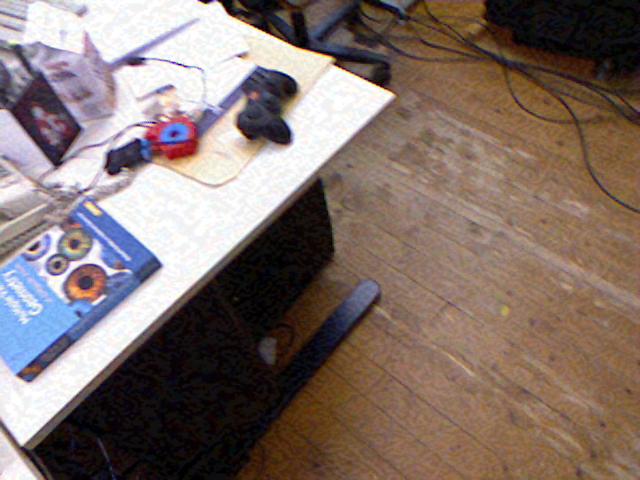}
         \caption{$\epsilon = 0.05$}
     \end{subfigure}
     \hfill
     \begin{subfigure}[b]{0.45\columnwidth}
         \centering
         \includegraphics[width=\columnwidth]{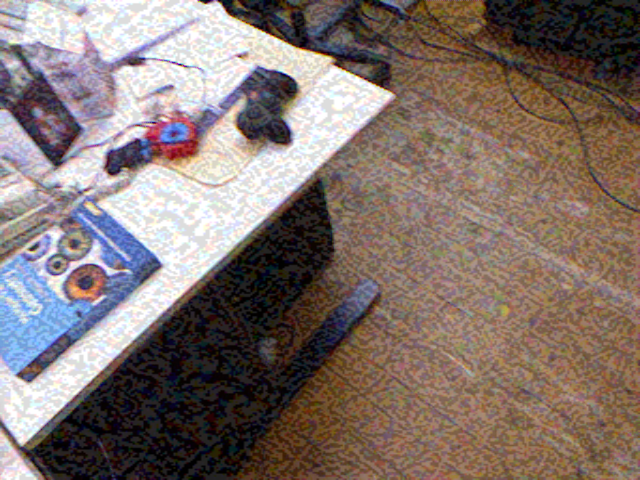}
         \caption{$\epsilon = 0.15$}
     \end{subfigure}
        \caption{Adding the adversarial perturbation (b) generated by the FGSM attack on the InceptionResNetV2 to the (a) original \textit{fr1\_360} image for (c) $\epsilon=0.05$ and (d) $\epsilon=0.15$.}
        \label{fig:epsilon}
\end{figure*}

\subsubsection{FGSM (Fast Gradient Sign Method)} FGSM computes the gradient of the loss function with respect to the input data and then applies a small perturbation in the direction of the sign of this gradient to create the adversarial example. FGSM is a single-step technique that is computationally efficient but may generate less potent adversarial examples compared to iterative methods. 
The adversarial example is calculated as:

\begin{equation}
    \begin{aligned}
        &x_{adv} = x + \delta \\
        &\delta = \epsilon \cdot sign(\nabla_xJ(x,y_{true}))
    \end{aligned}
\end{equation}

where $\delta$ is the FGSM perturbation, $\nabla_xJ(x,y_{true})$ corresponds to the gradient of the model loss function $J$ with respect to the input data $x$ for the true class $y_{true}$, the $sign$ function captures the sign of this gradient and $\epsilon$ is a small scalar value that determines the perturbation magnitude. Smaller $\epsilon$ values produce less noticeable perturbations, potentially resulting in less effective attacks. Larger $\epsilon$ values yield more prominent perturbations but might also be easier to detect. 
Fig.~\ref{fig:epsilon} illustrates the adversarial perturbations applied to an original image from the \textit{fr1\_360} trajectory, alongside the corresponding adversarial examples generated for various $\epsilon$ values.

\begin{figure*}[t]
     \centering
     \begin{subfigure}[b]{0.32\textwidth}
         \centering
         \includegraphics[width=\textwidth]{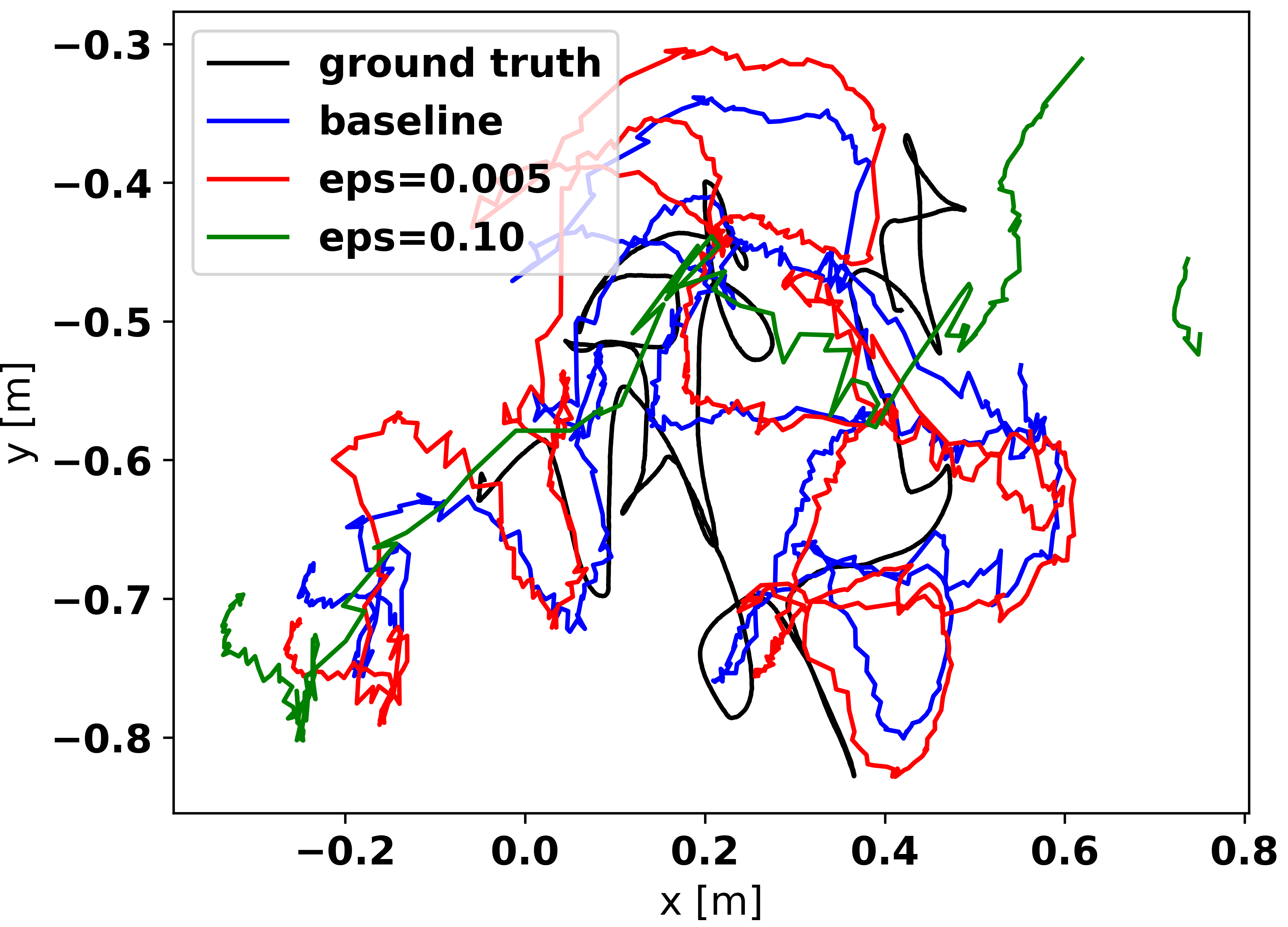}
         \caption{\textit{fr1\_360}}
         \label{fig:tr360}
     \end{subfigure}
     \hfill
     \begin{subfigure}[b]{0.32\textwidth}
         \centering
         \includegraphics[width=\textwidth]{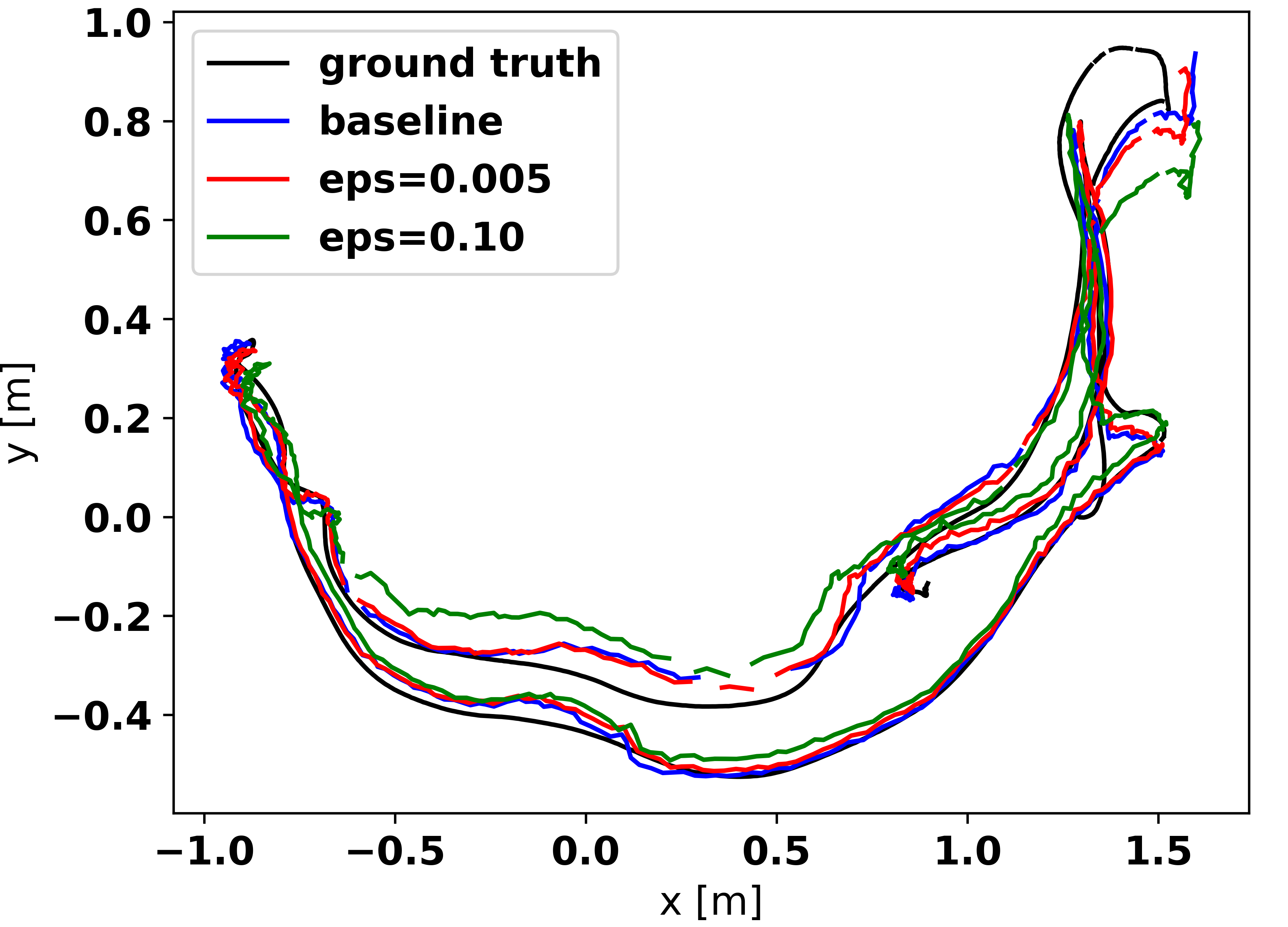}
         \caption{\textit{fr1\_desk2}}
         \label{fig:trdesk2}
     \end{subfigure}
     \hfill
     \begin{subfigure}[b]{0.32\textwidth}
         \centering
         \includegraphics[width=\textwidth]{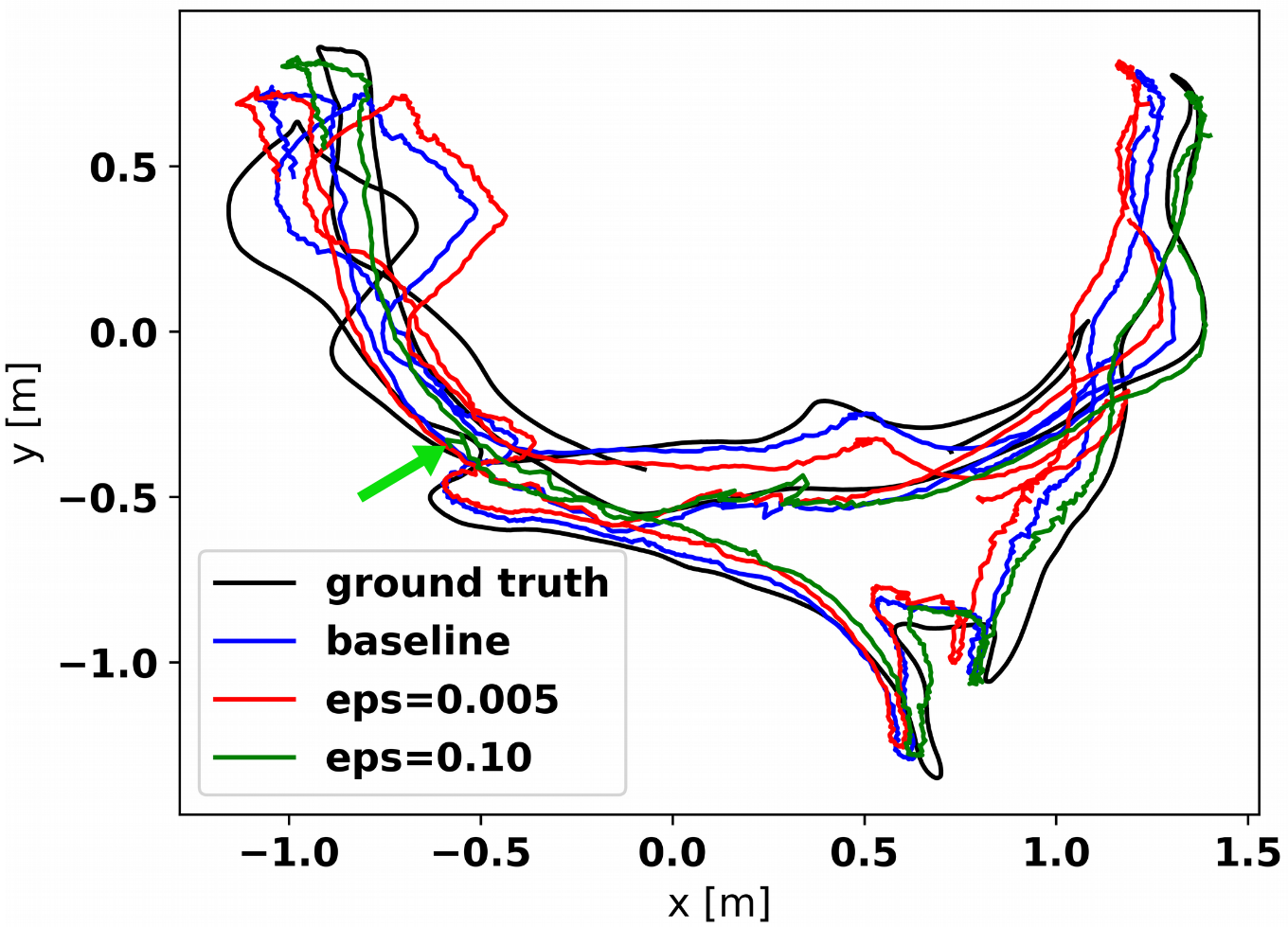}
         \caption{\textit{fr1\_room}}
         \label{fig:trroom}
     \end{subfigure}
        \caption{Origin-aligned ground truth, baseline and \textit{All-frames} untargeted FGSM attack for $\epsilon\in\{0.005, 0.10\}$ for three TUM trajectories.}
        \label{fig:trajerror}
\end{figure*}

\subsubsection{PGD (Projected Gradient Descent)}
PGD can be viewed as an extension of FGSM with multiple iterations. It is also a gradient-based method that introduces controlled perturbations to the input data using the $\epsilon$ parameter and aims to fool the target network. Notably, PGD offers enhanced control over the attack's intensity by allowing adjustments of key parameters such as the number of iterations $k$ and the step size $a$. The perturbation $\delta$ in the PGD attack is calculated as follows:

\begin{equation}
    \delta = \delta + \alpha \cdot sign(\nabla_xJ(x,y_{true}))
\end{equation}

At the end of each of the $k$ iterations, it ensures that $\delta$ remains within the 
range $(-\epsilon, \epsilon)$. 






\section{Experimental Evaluation}
\label{sec:experiments}

\subsection{Experimental setup}

We evaluate the FGSM and PGD attacks on the SLAM pipeline using the RGB-D images of the TUM dataset~\cite{sturm2012benchmark}. The RGB-D images have a 640x480 resolution, captured at a video frame rate of 30Hz. The TUM dataset also includes, for each such image, the ground truth data. 

We use two output quality metrics, the average Absolute Trajectory Error (ATE) and the percentage of untracked frames. The ATE quantifies the difference between points of the ground truth trajectory and the 
one estimated by SLAM. The frames that could not be tracked by the SLAM tracker are labelled as untracked, meaning that SLAM fails to estimate a pose. To compute the ATE in the presence of an untracked frame, we assume its pose to be identical to that of the most recent successfully tracked frame.

We use the InceptionResNetV2 CNN~\cite{SzegedyIVA17_InceptionResnet},  pre-trained on the ImageNet dataset~\cite{deng2009imagenet} as the known network (Fig.~\ref{fig:blackbox_bd}). The GCNv2 feature detector is the target network, i.e. the black-box. The impact on the system remains consistent whether MobileNetV2 or DenseNet is utilized as white-box networks, and, therefore, we  only focus on the InceptionResNetV2. Attacks are implemented using the CleverHans library~\cite{papernot2016technical}.

\begin{table*}[t]
    \caption{Average ATE across all frames (in meters)}
    \label{tab:ate}
    \resizebox{0.876\linewidth}{!}{%
    \centering
    \begin{tabularx}{\linewidth}{ l || c || r r r r r || r r r r r || r r r r r}
        \cline{1-17} \addlinespace[0.1cm]
        \textbf{Traj.} & \textbf{Baseline} & \multicolumn{5}{c||}{\textbf{All-frames}}  & \multicolumn{5}{c||}{\textbf{Time-Adaptive}}  & \multicolumn{5}{c}{\textbf{Spatially-Adaptive}} \\ \addlinespace[0.1cm]
        \cline{3-17} \addlinespace[0.1cm]
        \textbf{name}&  & \textbf{0.005} & \textbf{0.05 }& \textbf{0.10 }& \textbf{0.15 }& \textbf{0.30 }&  \textbf{0.005} & \textbf{0.05 }& \textbf{0.10 }& \textbf{0.15 }& \textbf{0.30 }& \textbf{0.005} & \textbf{0.05 }& \textbf{0.10 }& \textbf{0.15 }& \textbf{0.30 }\\ 
        \addlinespace[0.1cm]
        \cline{1-17} 
        \addlinespace[0.1cm]

        \textit{fr1\_360}	&	0.21	&	0.26	&	0.30	&	0.39	&	0.50	&	0.72	&	0.28	&	0.28	&	0.34	&	0.35	&	0.49	&	0.23	&	0.19	&	0.21	&	0.34	&	0.27	\\
        \textit{fr1\_floor}	&	0.33	&	0.33	&	0.35	&	0.47	&	0.56	&	0.67	&	0.32	&	0.33	&	0.34	&	0.34	&	0.60	&	0.35	&	0.32	&	0.33	&	0.34	&	0.34	\\
        \textit{fr1\_desk}	&	0.02	&	0.02	&	0.03	&	0.04	&	0.10	&	0.14	&	0.03	&	0.03	&	0.03	&	0.05	&	0.08	&	0.02	&	0.03	&	0.03	&	0.02	&	0.04	\\
        \textit{fr1\_desk2}	&	0.04	&	0.04	&	0.06	&	0.07	&	0.07	&	0.97	&	0.03	&	0.05	&	0.06	&	0.05	&	0.86	&	0.96	&	0.04	&	0.05	&	0.04	&	0.87	\\
        \textit{fr1\_room}	&	0.16	&	0.21	&	0.52	&	0.63	&	0.63	&	0.67	&	0.25	&	0.38	&	0.61	&	0.65	&	0.64	&	0.19	&	0.50	&	0.62	&	0.61	&	0.60	\\
        
        \addlinespace[0.1cm]
        \cline{1-17}
        \addlinespace[0.2cm]
    \end{tabularx}}
\end{table*}

\begin{table*}[hbt]
    \caption{Percentage of Untracked Frames}
    \label{tab:uf}
    \resizebox{0.876\linewidth}{!}{%
    \centering
    \begin{tabularx}{\linewidth}{ l || c || r r r r r || r r r r r || r r r r r}
        \cline{1-17} \addlinespace[0.1cm]
        \textbf{Traj.} & \textbf{Baseline} & \multicolumn{5}{c||}{\textbf{All-frames}}  & \multicolumn{5}{c||}{\textbf{Time-Adaptive}}  & \multicolumn{5}{c}{\textbf{Spatially-Adaptive}} \\ \addlinespace[0.1cm]
        \cline{3-17} \addlinespace[0.1cm]
        \textbf{name} &  & \textbf{0.005} & \textbf{0.05 }& \textbf{0.10 }& \textbf{0.15 }& \textbf{0.30 }&  \textbf{0.005} & \textbf{0.05 }& \textbf{0.10 }& \textbf{0.15 }& \textbf{0.30 }& \textbf{0.005} & \textbf{0.05 }& \textbf{0.10 }& \textbf{0.15 }& \textbf{0.30 }\\ 
        \addlinespace[0.1cm]
        \cline{1-17} 
        \addlinespace[0.1cm]

        \textit{fr1\_360}	&	0.0	&	0.0	&	75.7	&	75.8	&	75.8	&	76.6	&	0.0	&	0.0	&	0.0	&	35.1	&	76.1	&	0.0	&	0.0	&	0.0	&	75.8	&	0.0	\\
        \textit{fr1\_floor}	&	14.0	&	14.2	&	14.7	&	26.8	&	31.9	&	45.2	&	13.9	&	14.2	&	14.8	&	19.1	&	29.7	&	15.2	&	13.9	&	14.3	&	14.9	&	14.8	\\
        \textit{fr1\_desk}	&	0.0	&	0.0	&	0.0	&	0.0	&	0.0	&	0.0	&	0.0	&	0.0	&	0.0	&	0.0	&	0.0	&	0.0	&	0.0	&	0.0	&	0.0	&	0.0	\\
        \textit{fr1\_desk2}	&	0.2	&	0.2	&	0.0	&	0.6	&	0.0	&	96.0	&	0.2	&	0.0	&	0.0	&	0.2	&	96.0	&	91.9	&	0.0	&	0.0	&	0.6	&	95.6	\\
        \textit{fr1\_room}	&	0.0	&	0.0	&	0.0	&	38.5	&	38.9	&	39.1	&	0.0	&	27.0	&	38.3	&	38.5	&	38.8	&	0.0	&	37.4	&	38.2	&	38.5	&	38.7	\\
        
        \addlinespace[0.1cm]
        \cline{1-17}
    \end{tabularx}}
\end{table*}

\subsection{Untargeted attacks}
In this section, we examine the impact of untargeted black-box attacks on the SLAM system. As the results of the FGSM and PGD attack methods are similar, we focus here on the FGSM attack only.
We present four untargeted black-box FGSM attacks. 
\begin{itemize}
    \item \textit{All-frames}: This attack varies the $\epsilon$ parameter (as defined in the FGSM equation) while attacking all frames of the trajectory. 
     \item \textit{Rate}: Reduces the number of frames subjected to the attack compared to \textit{All-frames}. 
     \item \textit{Time-Adaptive}: An attack is applied only when the processing time of the frame (running ORB-SLAM) exceeds a threshold. This idea is to attack frames for which SLAM encounters challenges in locating a viable posing which is observable as more iterations needed to converge and thus require higher execution time.
     \item \textit{Spatially-Adaptive}: The attack is applied only to the regions of a frame that contains detected objects. The rationale is that the objects of an image are strongly correlated with important features.
\end{itemize}

\subsubsection{All-frames}
This attack is applied to all frames in each trajectory for $\epsilon \in \{0.005, 0.05, 0.10, 0.15, 0.30\}$. As described in section \ref{sec:methodology}, the larger the $\epsilon$, the stronger the attack. Table~\ref{tab:ate} shows the average ATE and Table~\ref{tab:uf} gives the percentage of untracked frames for different values of $\epsilon$. ``Baseline" refers to the attack-free case, which may still suffer from erroneous estimations since SLAM is just an approximation of the ground truth. Results confirm that despite its robustness and adaptivity, the SLAM pipeline is vulnerable even to small perturbations. For example, \textit{fr1\_360} has 75\% untracked frames even at a low $\epsilon=0.05$, while \textit{fr1\_room} has 38.5\% untracked frames at $\epsilon=0.1$.
Across all datasets, the attacks performed with $\epsilon=0.1$ result in an increase of the ATE 
by at least $43\%$ compared 
to baseline. 
 
The trajectories of the ground truth, the baseline, and the \textit{All-frames} untargeted attacks are shown in Fig.~\ref{fig:trajerror} (the z-coordinate is suppressed for clarity). 
The discontinuity of some trajectory lines such as the green line ($\epsilon=0.1$) of Fig.~\ref{fig:tr360} is due to the presence of untracked frames. This demonstrates how an agent's pose can be drastically affected when more than 70\% of frames are untracked.
Similarly, in the case of the \textit{fr1\_room} trajectory, the green line terminates sooner as the algorithm faces challenges in estimating the pose for the final 521 frames (see small arrow in Fig.~\ref{fig:trroom}). In Figs. \ref{fig:trdesk2} and \ref{fig:trroom}, we note fewer differences between the baseline and attacked trajectories in comparison to the \textit{fr1\_360} trajectory. 
Nevertheless, the impact of the attack remains significant, as even a deviation of 10cm can have serious implications for the performance of a SLAM system in various application scenarios.

In \textit{fr1\_360}, \textit{fr1\_floor}, and \textit{fr1\_room}, the agent moves indoors capturing scenes from an entire room, providing diverse images. 
Most of the \textit{ fr1\_floor} scenes show the floor, where the absence of objects makes the SLAM algorithm more vulnerable to attacks (hence, the large number of untracked frames). On the contrary, \textit{fr1\_desk} and \textit{fr1\_desk2} are more robust due to the small movements around two desks and the presence of several objects that make the differences between successive images noticeable. In \textit{fr1\_desk2}, 
the faster camera movements may increase the risk of tracking failure because of the possible image blurriness and the larger displacements between consecutive frames. 


\subsubsection{Rate}
An extension to the previous line of experiments is to limit the adversarial attacks to a subset of input frames at a frame selection rate $F$ (\textit{All-Frames} attack corresponds to $F=1$).  We have tried different values for $F \in \{1,1/2,1/3,1/4,1/5,1/6,1/7\}$.

\begin{figure}[htb]
  \centering    
  \includegraphics[width=\columnwidth]{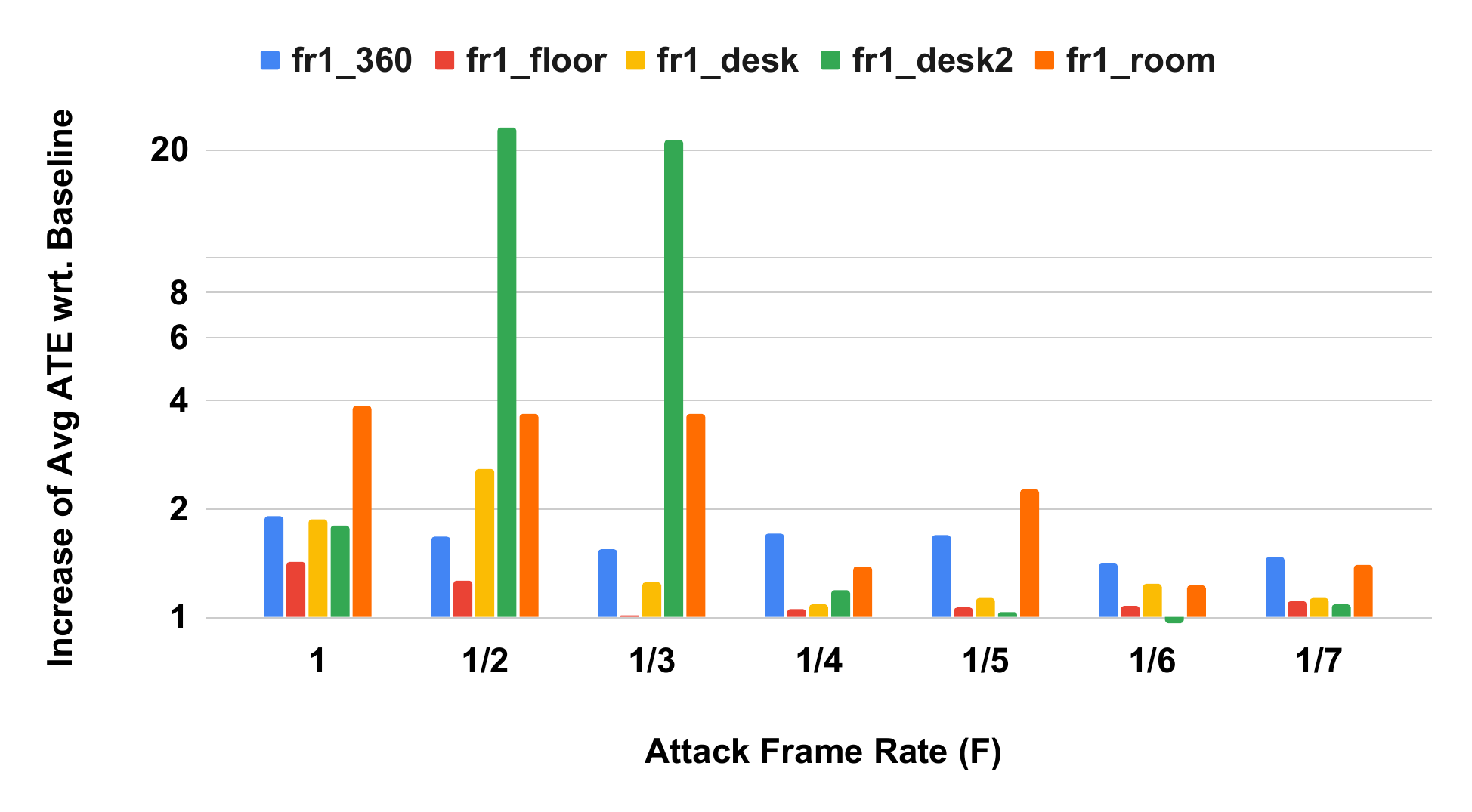}
  \caption{The increase of average ATE wrt. the baseline for each frame selection rate $F$  when $\epsilon=0.10$.}
  \label{fig:attackrate}
\end{figure} 

In Fig. \ref{fig:attackrate}, we observe that in many cases, ATE is comparable to, or even higher than when applying the \textit{All-frames} attack. However, the \textit{Rate} parameter is challenging to estimate due to the substantial influence of trajectory characteristics. 
Our experimentation with the \textit{Rate} attack aims to illustrate the randomness that can emerge due to the interaction of the attack with the frame selection mechanism applied by SLAM during the image-matching phase.
Notably, for the \textit{fr1\_desk2} trajectory, values of $F = 1/2$ and $F = 1/3$ severely disrupt the system, whereas \textit{fr1\_360} and \textit{fr1\_floor} remain unaffected for all values of $F < 1.$

\subsubsection{Time-Adaptive}
\begin{figure}[hbt]
  \centering    
  \includegraphics[width=\columnwidth]{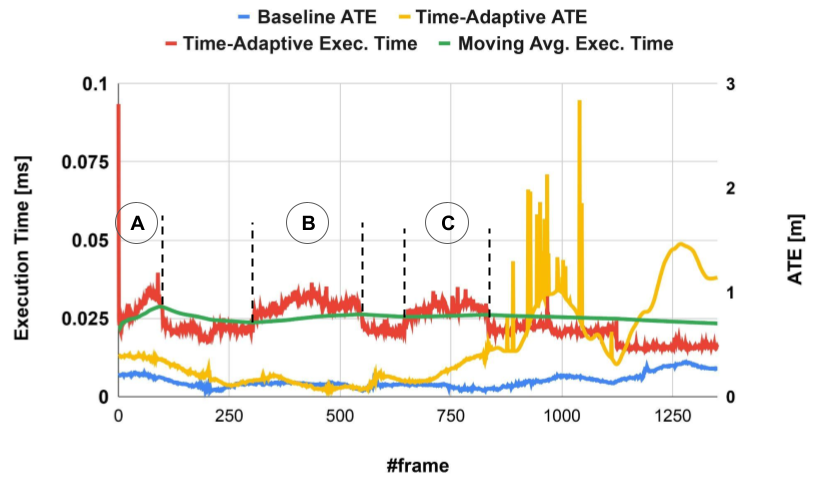}
  \caption{Timelines of execution time and ATE per frame of the \textit{fr1\_room} for the baseline and \textit{Time-Adaptive} attack when $\epsilon=0.10$. The attack is applied in each frame within the three regions A, B, and C. In these regions, SLAM execution time/frame is higher than the moving average of the execution time.}
  \label{fig:threshold}
\end{figure}

Tables \ref{tab:ate} and \ref{tab:uf} indicate that the ATE remains high when a frame is attacked at an adaptive rate. Fig.~\ref{fig:threshold} illustrates the two ATE timelines comparing the attacked trajectory (yellow line) with the baseline trajectory (blue line) for \textit{fr1\_room}. In this experiment, a frame $i$ is attacked with adversarial noise at $\epsilon=0.10$ every time the previous frame $i-1$ has an execution time (to perform SLAM) higher than the moving average of the execution time (green line). The red line indicates the SLAM execution time per frame. In this example, the average ATE is nearly equal to that of the \textit{All-frames} attack ($0.61$m compared to $0.63$m), even if only 39\% of the frames are attacked.

It is interesting to note the lag and also the cumulative impact of the adversarial attacks on the ATE. The ATE experiences a noticeable rise after frame 880, primarily due to the earlier regions A, B, and C, during which the input frame is subjected to an adversarial attack.



\subsubsection{Spatially-Adaptive}
\begin{figure}[hbt]
  \centering    
  \includegraphics[width=0.75\columnwidth]{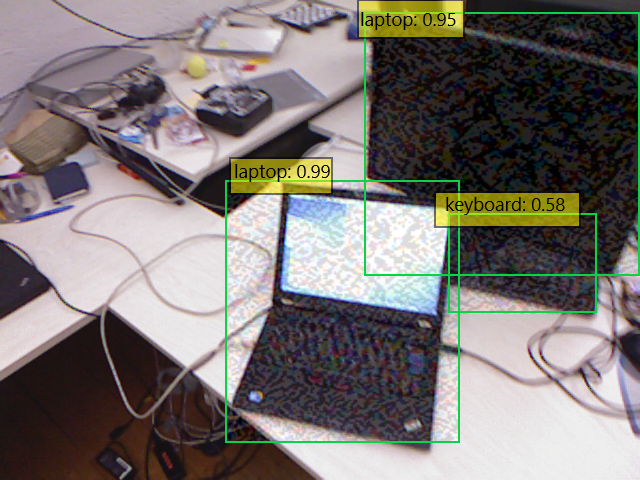}
  \caption{Application of the \textit{Spatially-Adaptive} attack for $\epsilon=0.15$. Only the pixels in the three regions are attacked.}
  \label{fig:yolonoise}
\end{figure}

The rightmost columns of Tables \ref{tab:ate} and \ref{tab:uf} show the results when the adversarial attack is applied to all the rectangular regions that have been detected by YOLOv4 to contain objects. 
We apply the attack on all detected objects of each frame. Fig.~\ref{fig:yolonoise} shows the noise on the image regions where YOLOv4 detects objects.
\textit{fr1\_room} exhibits comparable or even higher ATE values than the previous attacks. For instance, the increase in untracked frames occurs earlier compared to the \textit{All-frames} attack, when $\epsilon=0.05$. The impact on the system is less evident for the remaining trajectories. For example, when $\epsilon=0.30$, the \textit{All-frames} attack achieves almost $4x$ higher average ATE than the \textit{Spatially-Adaptive} for the \textit{fr1\_desk} trajectory. 
The error for the \textit{fr1\_floor} trajectory remains consistent regardless of the $\epsilon$ value. This stability is explained in Table~\ref{tab:percyolo} by the low percentage of frames, and the corresponding pixels within them, in which YOLOv4 detects objects. 

\begin{table}[htb]
\centering
\resizebox{\columnwidth}{!}{%
\begin{tabular}{|l|c|c|c|c|c|}
    \hline
    &  \multicolumn{5}{|c|}{\textbf{TUM datasets \textit{fr1\_*}}} \\
    \cline{2-6}
    & \textit{360} & \textit{floor} & \textit{desk} & \textit{desk2} & \textit{room}\\
    \hline
    \% frames & 59.4\% & 18.9\% & 97.7\% & 94.8\% & 88\% \\
    \% avg. pixels & 18.4\% & 12.4\% & 28.4\% & 29.2\% & 25.8\% \\
    \hline
\end{tabular}}
\caption{Percentage of frames and average percentage of attacked pixels in each frame in which YOLOv4 detects objects for each dataset.}
\label{tab:percyolo}
\end{table}

\subsubsection{Discussion}
\begin{figure}
    \centering
    \begin{subfigure}{\columnwidth}
        \begin{subfigure}{0.46\columnwidth}
            \includegraphics[width=\columnwidth]{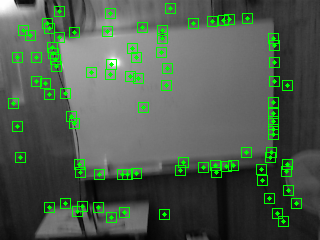}
        \end{subfigure}
        \hfill
        \begin{subfigure}{0.46\columnwidth}
            \includegraphics[width=\columnwidth]{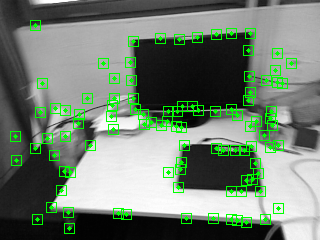}
        \end{subfigure}
        \caption{Original features}
        \label{fig:orfeat}
    \end{subfigure}
    \vfill
    \begin{subfigure}{\columnwidth}
        \begin{subfigure}{0.46\columnwidth}
            \includegraphics[width=\columnwidth]{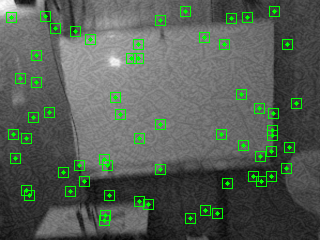}
        \end{subfigure}
        \hfill
        \begin{subfigure}{0.46\columnwidth}
            \includegraphics[width=\columnwidth]{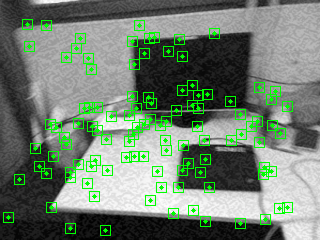}
        \end{subfigure}
        \caption{Attacked features}
        \label{fig:attfeat}
    \end{subfigure}
    \caption{The features detected by (a) the Original and (b) the Attacked implementation on frames of \textit{fr1\_360} (left column) and \textit{fr1\_room} (right column) trajectories.}
    \label{fig:features}
\end{figure}

The effectiveness of the adversarial attacks is explained by the redistribution of features compared to the initial implementation. The adversarial noise reduces the importance of the original features in the attacked images by distorting the pixels' color and quality and by creating clusters of noise in otherwise flat fields. 
Feature matching maintains pose tracking based on salient points appearing in consecutive images. In our experiments, the adversarial noise (which is unique in each frame) spreads these significant features to different positions, making the matching procedure less efficient. For example, in Fig.~\ref{fig:orfeat}, the original features are detected on the objects' outlines. The adversarial ones (in Fig.~\ref{fig:attfeat}) move irregularly into the neighboring area (in a different way for consecutive frames), increasing the false positive matches. Even if the matching algorithm finds actual similar points in the adversarial experiment, the error will gradually increase compared to the original implementation. 
The fact that significant features are located near objects validates the effectiveness of the \textit{Spatially-Adaptive} attack since it attacks the regions where objects are detected. 


\subsection{Targeted attacks}

Targeted black-box adversarial attacks are typically introduced through a process that involves several steps. Since FGSM involves only a single gradient step, it lacks the strength necessary to succeed as a targeted attack. Hence, we limit our experiments to employing the PGD attack method. The difference between images resulting from a targeted and untargeted attack is imperceptible. 

To generate the targeted adversarial input, we need a target label. Since we attack a feature detector, it is reasonable to anticipate that opting for distinct labels in successive frames introduces varied forms of noise, heightening the potential threat posed by the attack. We showed that assigning the same target label to all the images makes the attack less aggressive.
We therefore implement our attacks following the approach of Fig. \ref{fig:blackbox_bd} and randomly select the target label for each input image. 

\begin{figure}[bt]
  \centering    
  \includegraphics[width=\columnwidth]{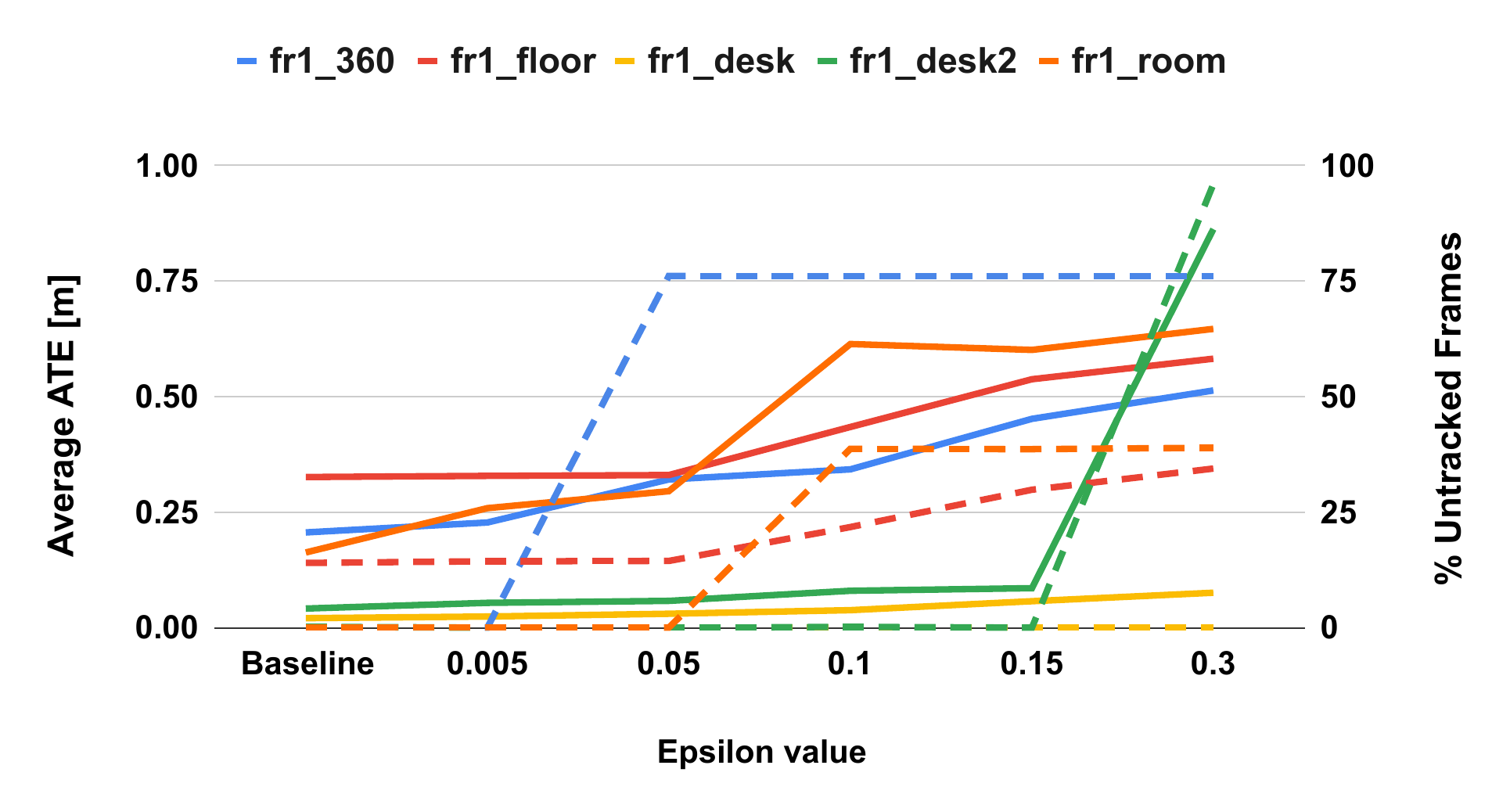}
  \caption{The effect of targeted black-box PGD attack on SLAM accuracy metrics. The dashed lines show the percentage of untracked frames for each trajectory (right axis).} 
  \label{fig:targeted}
\end{figure}

Fig.~\ref{fig:targeted} shows that in all trajectories the average ATE of the targeted attack is lower than the average ATE of the untargeted \textit{All-frames} attack for $\epsilon$ higher than $0.10$. 
When $\epsilon$ equals $0.005$, the targeted attack increases ATE by up to 30\%, compared to the untargeted \textit{All-frames} attack (Table~\ref{tab:ate}). 



\subsection{Attack of depth images}
SLAM input consists of RGB and Depth frames. Depth images are used for many of the stages of the ORB-SLAM2 back-end. SLAM calculates a virtual coordinate for each extracted feature using its depth value. The depth value also determines if a feature is used to provide translation or rotation information according to its distance from the camera. During SLAM localization, depth information from the previous frames forms 3D points which are used in matching with the current frame's features. Hence, the depth sensor's uncertainty plays a critical role in the SLAM system, a fact corroborated by our experimental findings.

When implementing our black-box attack on the depth images, the SLAM system consistently fails to estimate a single-frame pose. This holds true regardless of whether the noise is applied solely to the depth image or to both the depth and RGB images.


\subsection{Transferability}
\label{subsec:transferability}

In this section, we investigate whether adversarial attacks on feature detectors can be applied to alternative 
models with distinct architectures and loss functions.
We explored MobileNetV2 and DenseNet in addition to InceptionResNetV2 in the role of the known network of Fig.~\ref{fig:blackbox_bd}. Remarkably, the outcomes exhibited striking similarity across all three networks. Hence, for brevity, we have opted to present only the experiments conducted with InceptionResNetV2.

\begin{table}[hbt]
    \caption{Percentage of Untracked Frames for \textit{All-frames} attack on DXSLAM}
    \label{tab:dxslam}
    \resizebox{0.93\columnwidth}{!}{%
    \centering
    \begin{tabularx}{\columnwidth}{l | c | c c c c c}
        \cline{1-7} \addlinespace[0.1cm]
        \textbf{Traj.} & \textbf{Baseline} & \multicolumn{5}{c}{\textbf{$\epsilon$ value}} \\ 
        \textbf{name} &  & \textbf{0.005} & \textbf{0.05} & \textbf{0.10} & \textbf{0.15} & \textbf{0.30} \\ 
        \addlinespace[0.1cm]
        \cline{1-7} 
        \addlinespace[0.1cm]

        \textit{fr1\_360}	&	9.3	&	9.7	&	95.7	&	98.5	&	97.4	&	98.7	\\
        \textit{fr1\_floor}	&	12.8	&	13.4	&	35.6	&	39.9	&	82.8	&	100.0	\\
        \textit{fr1\_desk}	&	65.8	&	52.0	&	68.9	&	97.2	&	98.1	&	96.0	\\
        \textit{fr1\_room}	&	51.7	&	61.4	&	15.8	&	97.0	&	97.5	&	96.9	\\
        \cline{1-7}
    \end{tabularx}}
\end{table}

Since our attack is black-box and independent of the target network, it can be applied to any SLAM system. 
We apply our black-box untargeted \textit{All-frames} attack to the DXSLAM feature detector~\cite{li2020dxslam}, 
the only open-source system other than GCN-SLAM that combines a CNN-based feature detector with a SLAM back-end.
Table~\ref{tab:dxslam} gives the percentage of untracked frames for all the trajectories except the \textit{fr1\_desk2} trajectory, for which 
the baseline DXSLAM implementation fails to track $98\%$ of the frames. 
We note that untargeted adversarial perturbations, initially designed to target GCN-SLAM, can also be applied to attack DXSLAM, a model with a distinct architecture. In certain instances,
 the impact of the attack on DXSLAM is even more pronounced compared to the GCN-SLAM.

\section{Conclusions}
\label{sec:concl}

We evaluate black-box adversarial attacks on a CNN-based feature detector used as the front-end of a SLAM system. Our experiments with the TUM dataset reveal that a SLAM system is vulnerable to adversarial manipulation. 
These attacks, executed using methods like FGSM or PGD, negatively impact the accuracy of SLAM even when the perturbation weight $\epsilon$ is set to a low value or only a subset of frames is targeted.
 Moreover, our investigation underscores the severe repercussions of attacking depth images, which can be disastrous to the SLAM system.
This research shows that SLAM systems are susceptible to  adversarial attacks directed at their feature detectors and 
underscore the critical need for future research to prioritize the development of defense mechanisms, whether by strengthening algorithm robustness in the SLAM back-end or by implementing effective detection and protection mechanisms.
{
    \small
    \bibliographystyle{ieeenat_fullname}
    \bibliography{bibliography}

\begin{thebibliography}{38}
\providecommand{\natexlab}[1]{#1}
\providecommand{\url}[1]{\texttt{#1}}
\expandafter\ifx\csname urlstyle\endcsname\relax
  \providecommand{\doi}[1]{doi: #1}\else
  \providecommand{\doi}{doi: \begingroup \urlstyle{rm}\Url}\fi

\bibitem[Abdelfattah et~al.(2021)Abdelfattah, Yuan, Wang, and Ward]{abdelfattah2021adversarial}
Mazen Abdelfattah, Kaiwen Yuan, Z~Jane Wang, and Rabab Ward.
\newblock {Adversarial Attacks on Camera-Lidar Models for 3D Car Detection}.
\newblock In \emph{2021 IEEE/RSJ International Conference on Intelligent Robots and Systems (IROS)}, pages 2189--2194. IEEE, 2021.

\bibitem[Arnab et~al.(2018)Arnab, Miksik, and Torr]{ArnabMT18SemSegm}
Anurag Arnab, Ondrej Miksik, and Philip H.~S. Torr.
\newblock {On the Robustness of Semantic Segmentation Models to Adversarial Attacks}.
\newblock In \emph{{IEEE} Conference on Computer Vision and Pattern Recognition, {CVPR}, Salt Lake City, UT, USA, June 18-22, 2018}, pages 888--897, 2018.

\bibitem[Bochkovskiy et~al.(2020)Bochkovskiy, Wang, and Liao]{bochkovskiy2020yolov4}
Alexey Bochkovskiy, Chien-Yao Wang, and Hong-Yuan~Mark Liao.
\newblock {Yolov4: Optimal Speed and Accuracy of Object Detection}.
\newblock \emph{arXiv preprint arXiv:2004.10934}, 2020.

\bibitem[Carlini and Wagner(2017)]{carlini2017towards}
Nicholas Carlini and David Wagner.
\newblock {Towards Avaluating the Robustness of Neural Networks}.
\newblock In \emph{2017 ieee symposium on security and privacy (sp)}, pages 39--57. Ieee, 2017.

\bibitem[Chawla et~al.(2022)Chawla, Varma, Arani, and Zonooz]{chawla2022adversarial}
Hemang Chawla, Arnav Varma, Elahe Arani, and Bahram Zonooz.
\newblock {Adversarial Attacks on Monocular Pose Estimation}.
\newblock In \emph{2022 IEEE/RSJ International Conference on Intelligent Robots and Systems (IROS)}, pages 12500--12505. IEEE, 2022.

\bibitem[Chen et~al.()Chen, Luo, Zhou, Tian, Zhen, Fang, McKinnon, Tsin, and Quan]{ChenLZTZFMTQ22ASpanFormer}
Hongkai Chen, Zixin Luo, Lei Zhou, Yurun Tian, Mingmin Zhen, Tian Fang, David McKinnon, Yanghai Tsin, and Long Quan.
\newblock {ASpanFormer: Detector-Free Image Matching with Adaptive Span Transformer}.
\newblock In \emph{17th European Conference on Computer Vision (ECCV), Tel Aviv, Israel, October 23-27, 2022, Part {XXXII}}, pages 20--36.

\bibitem[Deng et~al.(2009)Deng, Dong, Socher, Li, Li, and Fei-Fei]{deng2009imagenet}
Jia Deng, Wei Dong, Richard Socher, Li-Jia Li, Kai Li, and Li Fei-Fei.
\newblock Imagenet: A large-scale hierarchical image database.
\newblock In \emph{2009 IEEE conference on computer vision and pattern recognition}, pages 248--255. Ieee, 2009.

\bibitem[DeTone et~al.(2018)DeTone, Malisiewicz, and Rabinovich]{detone2018superpoint}
Daniel DeTone, Tomasz Malisiewicz, and Andrew Rabinovich.
\newblock {Superpoint: Self-supervised Interest Point Detection and Description}.
\newblock In \emph{Proceedings of the IEEE conference on computer vision and pattern recognition workshops}, pages 224--236, 2018.

\bibitem[Finlayson et~al.(2018)Finlayson, Chung, Kohane, and Beam]{finlayson2018adversarial}
Samuel~G Finlayson, Hyung~Won Chung, Isaac~S Kohane, and Andrew~L Beam.
\newblock Adversarial attacks against medical deep learning systems.
\newblock \emph{arXiv preprint arXiv:1804.05296}, 2018.

\bibitem[Gao et~al.(2022)Gao, Yan, and Dong]{Gao2022BlackboxAA}
Jin Gao, Diqun Yan, and Mingyu Dong.
\newblock {Black-box Adversarial Attacks through Speech Distortion for Speech Emotion Recognition}.
\newblock \emph{EURASIP Journal on Audio, Speech, and Music Processing}, 2022:\penalty0 1--10, 2022.

\bibitem[Goodfellow et~al.(2014)Goodfellow, Shlens, and Szegedy]{fgsm2014goodfellow}
Ian~J Goodfellow, Jonathon Shlens, and Christian Szegedy.
\newblock {Explaining and Harnessing Adversarial Examples}.
\newblock \emph{arXiv preprint arXiv:1412.6572}, 2014.

\bibitem[Guo et~al.(2019)Guo, Gardner, You, Wilson, and Weinberger]{guo2019simple}
Chuan Guo, Jacob Gardner, Yurong You, Andrew~Gordon Wilson, and Kilian Weinberger.
\newblock {Simple black-box Adversarial Attacks}.
\newblock In \emph{International Conference on Machine Learning}, pages 2484--2493. PMLR, 2019.

\bibitem[Haoran et~al.(2021)Haoran, Yu-an, Yuan, Yajie, and Jingfeng]{Haoran2021ACA}
Lyu Haoran, Tan Yu-an, Xue Yuan, Wang Yajie, and Xu Jingfeng.
\newblock {A CMA‐ES‐Based Adversarial Attack Against Black‐Box Object Detectors}.
\newblock \emph{Chinese Journal of Electronics}, 30:\penalty0 406--412, 2021.

\bibitem[Huang et~al.(2017)Huang, Liu, Van Der~Maaten, and Weinberger]{huang2017densely}
Gao Huang, Zhuang Liu, Laurens Van Der~Maaten, and Kilian~Q Weinberger.
\newblock Densely connected convolutional networks.
\newblock In \emph{Proceedings of the IEEE conference on computer vision and pattern recognition}, pages 4700--4708, 2017.

\bibitem[Ilyas et~al.(2018)Ilyas, Engstrom, Athalye, and Lin]{ilyas2018black}
Andrew Ilyas, Logan Engstrom, Anish Athalye, and Jessy Lin.
\newblock {Black-box Adversarial Attacks with Limited Queries and Information}.
\newblock In \emph{International conference on machine learning}, pages 2137--2146. PMLR, 2018.

\bibitem[Im~Choi and Tian(2022)]{im2022adversarial}
Jung Im~Choi and Qing Tian.
\newblock {Adversarial Attack and Defense of Yolo Detectors in Autonomous Driving Scenarios}.
\newblock In \emph{2022 IEEE Intelligent Vehicles Symposium (IV)}, pages 1011--1017. IEEE, 2022.

\bibitem[Jiang et~al.(2019)Jiang, Ma, Chen, Bailey, and Jiang]{jiang2019black}
Linxi Jiang, Xingjun Ma, Shaoxiang Chen, James Bailey, and Yu-Gang Jiang.
\newblock {Black-box Adversarial Attacks on Video Recognition Models}.
\newblock In \emph{Proceedings of the 27th ACM International Conference on Multimedia}, pages 864--872, 2019.

\bibitem[Khan et~al.(2022)Khan, Naseer, Hayat, Zamir, Khan, and Shah]{KhanNHZKS22_VisionTransf}
Salman~H. Khan, Muzammal Naseer, Munawar Hayat, Syed~Waqas Zamir, Fahad~Shahbaz Khan, and Mubarak Shah.
\newblock {Transformers in Vision: {A} Survey}.
\newblock \emph{{ACM} Comput. Surv.}, 54\penalty0 (10s):\penalty0 200:1--200:41, 2022.

\bibitem[Lapid and Sipper(2023)]{lapid2023patch}
Raz Lapid and Moshe Sipper.
\newblock {Patch of Invisibility: Naturalistic Black-box Adversarial Attacks on Object Detectors}.
\newblock \emph{arXiv preprint arXiv:2303.04238}, 2023.

\bibitem[Li et~al.(2020)Li, Shi, Long, Liu, Yang, Wang, Wei, and Qiao]{li2020dxslam}
Dongjiang Li, Xuesong Shi, Qiwei Long, Shenghui Liu, Wei Yang, Fangshi Wang, Qi Wei, and Fei Qiao.
\newblock {DXSLAM: A Robust and Efficient Visual SLAM System with Deep Features}.
\newblock In \emph{2020 IEEE/RSJ International conference on intelligent robots and systems (IROS)}, pages 4958--4965. IEEE, 2020.

\bibitem[Lindenberger et~al.(2023)Lindenberger, Sarlin, and Pollefeys]{LightGlue}
Philipp Lindenberger, Paul{-}Edouard Sarlin, and Marc Pollefeys.
\newblock {LightGlue: Local Feature Matching at Light Speed}.
\newblock \emph{CoRR}, abs/2306.13643, 2023.

\bibitem[Liu et~al.(2018)Liu, Yang, Liu, Song, Chen, and Li]{Liu2018DPATCHAA}
Xin Liu, Huanrui Yang, Ziwei Liu, Linghao Song, Yiran Chen, and Hai~Helen Li.
\newblock {DPATCH: An Adversarial Patch Attack on Object Detectors}.
\newblock \emph{arXiv: Computer Vision and Pattern Recognition}, 2018.

\bibitem[Madry et~al.(2017)Madry, Makelov, Schmidt, Tsipras, and Vladu]{pgd2017madry}
Aleksander Madry, Aleksandar Makelov, Ludwig Schmidt, Dimitris Tsipras, and Adrian Vladu.
\newblock {Towards Deep Learning Models resistant to Adversarial Attacks}.
\newblock \emph{arXiv preprint arXiv:1706.06083}, 2017.

\bibitem[Mur-Artal and Tard{\'o}s(2017)]{mur2017orb}
Raul Mur-Artal and Juan~D Tard{\'o}s.
\newblock {ORB-SLAM2: An Open-Source SLAM System for Monocular, Stereo, and RGB-D Cameras}.
\newblock \emph{IEEE transactions on robotics}, 33\penalty0 (5):\penalty0 1255--1262, 2017.

\bibitem[Nemcovsky et~al.(2022)Nemcovsky, Jacoby, Bronstein, and Baskin]{nemcovsky2022physical}
Yaniv Nemcovsky, Matan Jacoby, Alex~M Bronstein, and Chaim Baskin.
\newblock {Physical Passive Patch Adversarial Attacks on Visual Odometry Systems}.
\newblock In \emph{Proceedings of the Asian Conference on Computer Vision}, pages 1795--1811, 2022.

\bibitem[Papernot et~al.(2016)Papernot, Faghri, Carlini, Goodfellow, Feinman, Kurakin, Xie, Sharma, Brown, Roy, et~al.]{papernot2016technical}
Nicolas Papernot, Fartash Faghri, Nicholas Carlini, Ian Goodfellow, Reuben Feinman, Alexey Kurakin, Cihang Xie, Yash Sharma, Tom Brown, Aurko Roy, et~al.
\newblock {Technical Report on the Cleverhans v2. 1.0 Adversarial Examples Library}.
\newblock \emph{arXiv preprint arXiv:1610.00768}, 2016.

\bibitem[Papernot et~al.(2017)Papernot, McDaniel, Goodfellow, Jha, Celik, and Swami]{papernot2017practical}
Nicolas Papernot, Patrick McDaniel, Ian Goodfellow, Somesh Jha, Z~Berkay Celik, and Ananthram Swami.
\newblock {Practical Black-box attacks Against Machine Learning}.
\newblock In \emph{ACM Asia conference on computer and communications security}, pages 506--519, 2017.

\bibitem[Sandler et~al.(2018)Sandler, Howard, Zhu, Zhmoginov, and Chen]{sandler2018mobilenetv2}
Mark Sandler, Andrew Howard, Menglong Zhu, Andrey Zhmoginov, and Liang-Chieh Chen.
\newblock Mobilenetv2: Inverted residuals and linear bottlenecks.
\newblock In \emph{Proceedings of the IEEE conference on computer vision and pattern recognition}, pages 4510--4520, 2018.

\bibitem[Sarlin et~al.(2020)Sarlin, DeTone, Malisiewicz, and Rabinovich]{SarlinDMR20}
Paul{-}Edouard Sarlin, Daniel DeTone, Tomasz Malisiewicz, and Andrew Rabinovich.
\newblock {SuperGlue: Learning Feature Matching With Graph Neural Networks}.
\newblock In \emph{2020 {IEEE/CVF} Conference on Computer Vision and Pattern Recognition, {CVPR} 2020, Seattle, WA, USA, June 13-19, 2020}, pages 4937--4946. Computer Vision Foundation / {IEEE}, 2020.

\bibitem[Sturm et~al.(2012)Sturm, Engelhard, Endres, Burgard, and Cremers]{sturm2012benchmark}
J{\"u}rgen Sturm, Nikolas Engelhard, Felix Endres, Wolfram Burgard, and Daniel Cremers.
\newblock {A Benchmark for the Evaluation of RGB-D SLAM Systems}.
\newblock In \emph{2012 IEEE/RSJ international conference on intelligent robots and systems}, pages 573--580. IEEE, 2012.

\bibitem[Sun et~al.(2021)Sun, Shen, Wang, Bao, and Zhou]{sun2021loftr}
Jiaming Sun, Zehong Shen, Yuang Wang, Hujun Bao, and Xiaowei Zhou.
\newblock Loftr: Detector-free local feature matching with transformers.
\newblock In \emph{Proceedings of the IEEE/CVF conference on computer vision and pattern recognition}, pages 8922--8931, 2021.

\bibitem[Szegedy et~al.(2017)Szegedy, Ioffe, Vanhoucke, and Alemi]{SzegedyIVA17_InceptionResnet}
Christian Szegedy, Sergey Ioffe, Vincent Vanhoucke, and Alexander~A. Alemi.
\newblock {Inception-v4, Inception-ResNet and the Impact of Residual Connections on Learning}.
\newblock In \emph{{Thirty-First {AAAI} Conference on Artificial Intelligence, February 4-9, 2017, San Francisco, California, USA}}, pages 4278--4284, 2017.

\bibitem[Tang et~al.(2019)Tang, Ericson, Folkesson, and Jensfelt]{tang2019gcnv2}
Jiexiong Tang, Ludvig Ericson, John Folkesson, and Patric Jensfelt.
\newblock {GCNv2: Efficient Correspondence Prediction for Real-Time SLAM}.
\newblock \emph{IEEE Robotics and Automation Letters}, 4\penalty0 (4):\penalty0 3505--3512, 2019.

\bibitem[Wang et~al.(2022{\natexlab{a}})Wang, Zhang, Yang, Peng, and Stiefelhagen]{WangZYPS22Matchformer}
Qing Wang, Jiaming Zhang, Kailun Yang, Kunyu Peng, and Rainer Stiefelhagen.
\newblock {MatchFormer: Interleaving Attention in Transformers for Feature Matching}.
\newblock In \emph{16th Asian Conference on Computer Vision (ACCV), Macao, China, December 4-8, 2022, Part {III}}, pages 256--273, 2022{\natexlab{a}}.

\bibitem[Wang et~al.(2022{\natexlab{b}})Wang, Liu, Chang, Rodr{\'{\i}}guez, and Wang]{WangLCRW22WhiteBoxAttack}
Yixiang Wang, Jiqiang Liu, Xiaolin Chang, Ricardo~J. Rodr{\'{\i}}guez, and Jianhua Wang.
\newblock {DI-AA: An interpretable White-box Attack for Fooling Deep Neural Networks}.
\newblock \emph{Inf. Sci.}, 610:\penalty0 14--32, 2022{\natexlab{b}}.

\bibitem[Xie et~al.(2021)Xie, Wang, Kong, and Hong]{Xie2021Universal3P}
Shangyu Xie, Han Wang, Yu Kong, and Yuan Hong.
\newblock {Universal 3-Dimensional Perturbations for Black-Box Attacks on Video Recognition Systems}.
\newblock \emph{2022 IEEE Symposium on Security and Privacy (SP)}, pages 1390--1407, 2021.

\bibitem[Yao et~al.(2020)Yao, He, Han, and Zhou]{yao2020miss}
Qingsong Yao, Zecheng He, Hu Han, and S~Kevin Zhou.
\newblock {Miss the point: Targeted adversarial Attack on Multiple Landmark Detection}.
\newblock In \emph{Medical Image Computing and Computer Assisted Intervention--MICCAI 2020: 23rd International Conference, Lima, Peru, October 4--8, 2020}, pages 692--702, 2020.

\bibitem[Yuan et~al.(2019)Yuan, He, Zhu, and Li]{YuanHZL19AdvExamples}
Xiaoyong Yuan, Pan He, Qile Zhu, and Xiaolin Li.
\newblock {Adversarial Examples: Attacks and Defenses for Deep Learning}.
\newblock \emph{{IEEE} Trans. Neural Networks Learn. Syst.}, 30\penalty0 (9):\penalty0 2805--2824, 2019.

\end{thebibliography}
}


\end{document}